\pdfoutput=1

\documentclass[11pt]{article}
\usepackage[table]{xcolor} 
\usepackage[final]{acl}
\usepackage{times}
\usepackage{latexsym}
\usepackage[T1]{fontenc}
\usepackage{inconsolata}
\usepackage{nicefrac}       
\usepackage{soul}
\usepackage{url}
\usepackage[utf8]{inputenc}
\usepackage{bm}
\usepackage{amsmath}
\usepackage{amsthm}
\usepackage{booktabs}
\usepackage{algorithm}
\usepackage{enumitem}
\usepackage{mathrsfs}
\usepackage{cuted} 
\newcommand{\nsstitle}[1]{\noindent\textup{\textbf{#1}}}

\usepackage{microtype}
\usepackage{subfigure}
\usepackage{booktabs} 
\usepackage[many]{tcolorbox}    	
\usepackage{setspace}               
\usepackage{arydshln}
\definecolor{black}{HTML}{000000}  
\definecolor{blue_sub}{HTML}{cde4ff}     
\tcbset{
    sharp corners,
    colback = white,
    before skip = 0.2cm,    
    after skip = 0.5cm      
}  
\newtcolorbox{boxH}{
    colback = sub, 
    colframe = main, 
    boxrule = 0pt, 
    leftrule = 6pt 
}
\newtcolorbox{boxE}{
    rounded corners,
    arc = 3pt,
    colframe = black!50, 
    enhanced, 
    boxrule =0.2mm, 
    borderline = {0.2mm}{0pt}{black!20}, 
    left=1mm, 
    right=1mm, 
    top=1mm, 
    bottom=1mm 
}
\usepackage{tikz}

\usepackage{amssymb}
\usepackage{amsmath}
\usepackage{amsthm}
\usepackage{multirow}
\usepackage{makecell}
\def\shorten{\looseness=-1} 

\AtBeginDocument{%
  \providecommand\BibTeX{{%
    \normalfont B\kern-0.5em{\scshape i\kern-0.25em b}\kern-0.8em\TeX}}}
\usepackage{algpseudocode} 
\algdef{SE}[SUBALG]{Indent}{EndIndent}{}{\algorithmicend\ }%
\algtext*{Indent}
\algtext*{EndIndent}

\newcommand{\sysName}{GLOW}
\newcommand{\sysNameV}{GLOW}
\newcommand{\Taxonomy}{GLOW-Bench}

\author{
Hussein Abdallah\textsuperscript{1},
Ibrahim Abdelaziz\textsuperscript{2},
Panos Kalnis\textsuperscript{3}\and
Essam Mansour\textsuperscript{1}\\
\textsuperscript{1}Concordia University,
\textsuperscript{2}IBM,
\textsuperscript{3}KAUST\\
hussein.abdallah@mail.concordia.ca,
ibrahim.abdelaziz1@ibm.com,\\
panos.kalnis@kaust.edu.sa,
essam.mansour@concordia.ca
}
\title{Leveraging LLM-GNN Integration for Open-World Question Answering over Knowledge Graphs}
\begin{document}

\maketitle
\begin{abstract}
Open-world Question Answering (OW-QA) over knowledge graphs (KGs) aims to answer questions over incomplete or evolving KGs.
Traditional KGQA assumes a closed world where answers must exist in the KG, limiting real-world applicability. In contrast, open-world QA requires inferring missing knowledge based on graph structure and context.
Large language models (LLMs) excel at language understanding but lack structured reasoning. Graph neural networks (GNNs) model graph topology but struggle with semantic interpretation. Existing systems integrate LLMs with GNNs or graph retrievers. Some support open-world QA but rely on structural embeddings without semantic grounding. Most assume observed paths or complete graphs, making them unreliable under missing links or multi-hop reasoning.
We present \sysName{}, a hybrid system that combines a pre-trained GNN and an LLM for open-world KGQA. The GNN predicts top-$k$ candidate answers from the graph structure. These, along with relevant KG facts, are serialized into a structured prompt (e.g., triples and candidates) to guide the LLM's reasoning. This enables joint reasoning over symbolic and semantic signals, without relying on retrieval or fine-tuning. To evaluate generalization, we introduce \textsc{GLOW-Bench}, a 1{,}000-question benchmark over incomplete KGs across diverse domains. \sysName{} outperforms existing LLM–GNN systems on standard benchmarks and \textsc{GLOW-Bench}, achieving up to 53.3\% and an average 38\% improvement. GitHub code and data are available \href{https://github.com/CoDS-GCS/GLOW}{here}.
\end{abstract}

\section{Introduction}
\label{sec:intro}

Open-World Question Answering (OW-QA) over knowledge graphs (KGs) aims to answer questions when relevant facts are missing or the KG is incomplete. This challenge arises in real-world domains like biomedicine, scientific research, and finance, where knowledge is often evolving, implicit, or incomplete~\cite{COW_KGQA_Limitations,BioRAG}. For example, systems may need to infer undocumented drug interactions, latent collaborations, or causal relationships. In such cases, answers are not explicitly stored in the KG and must be predicted based on graph structure, entity semantics, and cues in the question.

Traditional KGQA methods assume a \emph{closed-world} setting, where all facts are known and retrievable. This limits their use in dynamic or incomplete environments. Open-world QA instead requires reasoning over both observed and missing information. Unlike standard retrieval or link prediction, it must integrate symbolic language with structural graph signals. This motivates hybrid approaches that combine the semantic flexibility of Large Language Models (LLMs) with the relational reasoning of Graph Neural Networks (GNNs).

Most existing KGQA systems rely on structured or dense retrieval to find explicit paths between question and answer entities. Methods like G-Retriever~\cite{G-Retriever}, GNN-RAG~\cite{GNNRAG}, RoG~\cite{RoG}, and ToG~\cite{ToG} retrieve graph fragments using semantic similarity. GCR~\cite{GCR} improves scalability but still assumes a complete graph. These methods often fail under open-world conditions, where answer paths are missing or poorly aligned with the question. This results in noisy retrieval and low accuracy. OW-QA instead requires \emph{predictive reasoning} to infer plausible answers beyond observed facts. AskGNN~\cite{Ask-GNN} addresses this by using GNNs to guide LLM inference, but relies heavily on structural embeddings and lacks semantic grounding. This limits its ability to handle complex or multi-hop questions.\shorten

\begin{table}[!t]
  \centering
  \caption{
Average accuracy (\%) on OW-QA benchmarks by reasoning depth. {\sysNameV}-GN leads on both 1- and 2-hop questions, while GCR drops sharply under OWA. All use Qwen3-8B; scores for existing datasets are averaged over arxiv2023, ogbn-arxiv, and ogbn-products.
}
  \label{tab:accuracy_kg_hops_grouped}
  \resizebox{0.9\columnwidth}{!}{%
  \begin{tabular}{l|c|cc}
    \toprule
    \multirow{2}{*}{\textbf{Method}} &\makecell{Existing\\Datasets}& \multicolumn{2}{c}{\makecell{{\Taxonomy} (ours)}} \\
    \cline{2-4}
     &\textbf{1-Hop} &\textbf{1-Hop} & \textbf{2-Hop} \\
    \hline
    LLM$_{\text{Only}}$& 25.7  & 30.0  & 18.9 \\
    GCR             &7.8   & 34.1&15.3  \\
     GoG             &  44.6 & 48.3  & 29.5 \\
    AskGNN             &  53.7 & 79.4  & 34.3 \\
    \cdashline{1-4}
    {\sysNameV}-GN     & \textbf{71.7}  & \textbf{83.3} & \textbf{42.4} \\
    \bottomrule
  \end{tabular} }
\end{table}

To address this challenge, we propose {\sysName}, a novel hybrid approach that synergistically combines the strengths of both LLMs and GNNs. Our method uses a pre-trained GNN to predict the top-k possible answers based on the graph topology. These answers, along with a serialized subgraph of KG facts relevant to the question, are then incorporated into the LLM prompt using a controlled and structured format (e.g., relational triples, top-ranked candidate answers), enabling the LLM to reason jointly over language and structure. Rather than relying on retrieval or pretraining alone, our model dynamically bridges gaps in the KG by augmenting the LLM with on-the-fly, GNN-driven prompts.\shorten 

We evaluate \sysName{} on the AskGNN~\cite{Ask-GNN} open-world QA benchmark, which is based on ogbn-arxiv, ogbn-products, and arxiv2023 datasets. AskGNN used these datasets to reflect realistic KG incompleteness and avoid the closed-world assumption in existing benchmarks, such as WebQSP or CWQ~\cite{GCR}. To test generalization, we introduce \textsc{\Taxonomy}, a new benchmark of 1{,}000 natural language questions across diverse domains. Each question requires reasoning over incomplete KGs, with the correct answer deliberately removed. 
Unlike AskGNN’s dataset, which covers only single-hop questions in less diverse domains, \textsc{\Taxonomy} includes 1- and 2-hop questions across multiple KGs.

We compare our method against state-of-the-art baselines, including AskGNN~\cite{Ask-GNN}, GCR~\cite{GCR}, a KG-grounded QA pipeline, GoG \cite{Generate-on-Graph},KGQA over incomplete KG, and LLMs such as GPT-4o-mini~\cite{gpt40mini} and Qwen3-8B~\cite{qwen3-8b}. As shown in Table~\ref{tab:accuracy_kg_hops_grouped}, LLMs perform poorly on deeper reasoning tasks, often failing to retrieve or organize relevant knowledge. AskGNN performs better but is limited by its structural focus and lack of semantic flexibility. GCR struggles under incomplete KGs, often hallucinating answer paths or returning incorrect answers.  GoG hallucinates answer path generation as it overlooks the underlying KG schema and structural constraints, resulting in semantically inconsistent or invalid paths. Our benchmark highlights these issues and shows the need for models that can reason jointly over symbolic and structural signals.
In summary, our contributions are:
\begin{itemize}[leftmargin=1.2mm]
\itemsep0em
    \item We propose \sysName{}, a novel OW-KGQA system that combines a GNN and an LLM to jointly reason over structured and unstructured knowledge.
    \item We present \textsc{\Taxonomy}, a 1,000-question benchmark for open-world KGQA with multi-hop reasoning over incomplete, cross-domain KGs. \shorten
    \item We demonstrate that \sysName{} outperforms state-of-the-art LLM–GNN QA and KGQA systems across standard benchmarks and \textsc{GLOW-Bench}, with up to 53.3\% and an average of 38\% improvement.\shorten
\end{itemize}


\section{Related Work}\label{sec:RW}

Our work connects to research in KG completion, LLM-based QA, GNN–LLM hybrid models, and open-world KGQA.

\textbf{Knowledge Graph Completion.}
Embedding-based methods (TransE, RotatE, ComplEx)~\cite{KGE_Survey} and recent semantic models (StructurE, HopfE, DensE)~\cite{KGE-Survey-2025} infer missing links via latent representations. Rule-based systems (AnyBURL~\cite{AnyBURL}, SAFRAN~\cite{SAFRAN}) generalize KG patterns with logical rules. While useful for KG augmentation, these approaches operate independently of QA. CBR-iKB applies case-based reasoning with KGEs but is computationally expensive and limited to transductive settings. In contrast, \sysName{} integrates KG completion into QA through GNN-guided prompting.

\textbf{LLMs for KGQA.}
Fine-tuning LLMs on KG triples can improve domain adaptation~\cite{StAR, KG-LLM, KG-FIT}, but requires extensive training and risks catastrophic forgetting~\cite{LLM_finetunning_cost, finetunning_forgetting}. Our approach avoids fine-tuning by injecting both graph-structured data and GNN outputs into LLM prompts, enabling semantic and structural reasoning without modifying LLM weights.

\textbf{GNN–LLM Hybrid Models.}
GNN-based QA systems~\cite{QA-GNN, ANet, KGTOSA} support multi-hop reasoning, but often ignore language semantics. Retrieval-augmented methods, such as GNN-RAG~\cite{GNNRAG}, G-Retriever~\cite{G-Retriever}, STaRK~\cite{STaRK}, and RoG~\cite{RoG}, embed graph fragments for LLMs to reason over. However, they assume complete graphs and rely on retrieving full answer paths, which breaks under missing links~\cite{QA_incomplete_KG}. Our method differs by using a GNN to predict candidate answers and relevant subgraphs, which are serialized into prompts for LLM reasoning.\shorten

\textbf{Open-World QA with In-Context Learning.}
AskGNN~\cite{Ask-GNN} enhances retrieval using GNN-based Structure-Enhanced Retrieval (SE-Retriever) to select in-context examples. However, it relies on joint LLM–GNN training using open-weight LLMs and scales poorly with the model and graph size. It may also bias predictions toward dominant classes. \sysName{} avoids these issues by using lightweight GNNs for candidate generation and prompt construction, without requiring fine-tuning, and generalizing to various LLMs. GoG~\cite{Generate-on-Graph} addresses KGQA over incomplete KGs by deliberately removing randomly answer path predicates. While effective, this approach cannot guarantee the complete elimination of all answer paths and their associated edges, and generates on the fly LLM-based triples that do not conform to the KG schema, hence causing answer hallucination.

\textbf{Dense KGQA via Path Retrieval.}
Methods like G-Retriever, GNN-RAG, ToG~\cite{ToG}, and RoG~\cite{RoG} retrieve semantically similar paths, assuming the answer exists in the KG. GCR~\cite{GCR} improves on RoG by fine-tuning LLMs to extract answers from retrieved paths. However, these systems fail in incomplete KG settings where the answer path is missing from the KG~\cite{QA_incomplete_KG}. Unlike them, \sysName{} supports predictive reasoning by prompting the LLM with GNN-predicted candidates and structured facts, even when no complete path exists.\shorten

\textbf{Benchmarks.}
Existing KGQA benchmarks (e.g., WebQuestionsSP~\cite{WebQuestionsSP}, LC-QuAD~\cite{Lc-quad}, MetaQA~\cite{MetaQA}, STaRK~\cite{STaRK}) assume closed-world settings with guaranteed answer paths. AskGNN~\cite{Ask-GNN} introduced open-world benchmarks but is limited to single-hop reasoning in narrow domains. We introduce \textsc{\Taxonomy}, a benchmark of 1{,}000 open-world questions requiring single- and multi-hop reasoning across diverse KGs, where gold answers are explicitly removed to test generalization under incompleteness.

\textbf{OW-KGQA vs. GNN Node Classification.}
Our task fundamentally differs from GNN node classification (NC); it takes natural language questions as input and infers answer nodes using KG structure and semantics, whereas NC operates on graph inputs with fixed labels and no linguistic reasoning. \shorten

\section{The \sysName{} Approach}
\label{sec:approach} 

\sysName{}\footnote{\textbf{G}raph-\textbf{L}LM for \textbf{O}pen-\textbf{W}orld QA} is a hybrid approach for open-world QA on KGs, combining graph-based reasoning with LLMs. Beyond \emph{In-Context Learning} (ICL), \sysName{} introduces \textit{In-Structure Learning}, where textual and KG signals jointly guide reasoning. A pre-trained GNN predicts top-$k$ candidates and retrieves a relevant KG subgraph based on the question’s entity. These are serialized into a structured prompt with relational triples and GNN predictions, enabling the LLM to reason over linguistic and structural cues without additional fine-tuning.

\begin{figure*}[!t]
  \centering
  \includegraphics[width=0.90\linewidth]{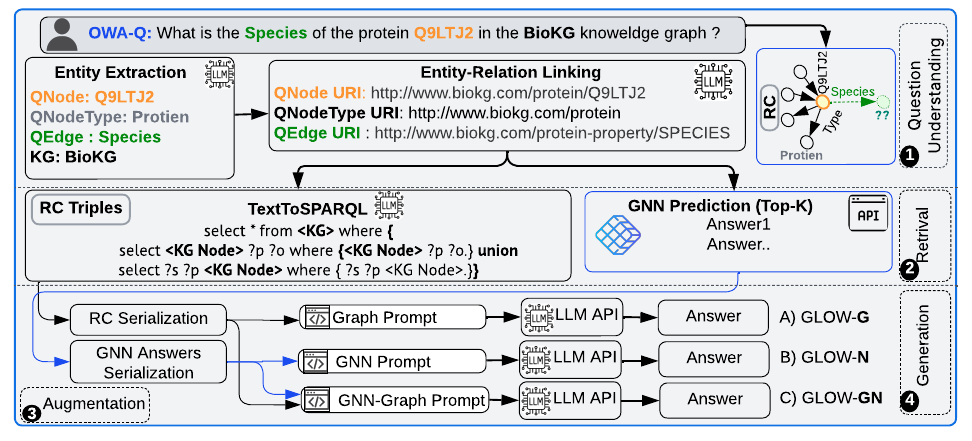}
  \caption{Overview of \sysName{}'s four stages: question understanding, retrieval, augmentation, and generation. \sysName{} builds GNN-guided prompts using KG context and/or GNN predictions, with three variants: (a) KG facts (GLOW-G), (b) GNN predictions (GLOW-N), and (c) both (GLOW-GN).}
  \label{fig:GLOWSys}
\end{figure*}

\subsection{\sysName{} Pipeline Overview}
\label{sec:glowqa_pipeline}

We develop three \sysName{} variants to explore different ways of integrating structured signals: \sysNameV{}-G (graph context), \sysNameV{}-N (GNN predictions), and \sysNameV{}-GN (combined). The full architecture is shown in Figure~\ref{fig:GLOWSys} and Algorithm ~\ref{algo:GLOW}. The pipeline proceeds in four stages: Question Understanding \& Linking, Retrieval, Augmentation, and Generation. Each contributes build a GNN-guided LLM prompt.\shorten

\begin{algorithm}[!t]
\small

\caption{\textsc{getPrompt}: Generate GLOW Prompt from Input OWA Question over KG}
\begin{algorithmic}[1]
\Require $Q$:User input question, $GLOW_v$: System variation
\Function{getPrompt}{$Q$,$GLOW_v$}
    \State $Q_n,Q_e,KG \gets \text{entityExtraction}(Q)$ \Comment{extract the question's node and edge}
    \State $KG_{Sc} \gets \text{getKGSchema}(KG)$
    \Comment{Load KG schema}
    \State $v_t,e_t\gets \text{ER-Linking}(Q,KG_{Sc},Q_n,Q_e)$
    \Comment{Link the question's node and edge crossponding KG's node and edge URIs}
    \State $\mathcal{RC}_q \gets \text{TextToSPARQL}(KG_{Sc}, v_t)$
    \Comment{generate $\mathcal{RC}$ equivalent SPARQL query}
    \State $\mathcal{RC}_{Triples} \gets \text{execSPARQL}
    (\mathcal{RC}_q,KG)$
    \Comment{Execute the $\mathcal{RC}$ SPARQL Query}
    \State $L \gets \text{getPossibleLabels}(v_t,e_t)$ 
    \Comment{Extract the set of possible labels}
    \State $RC \gets \text{RC-Serialization}(\mathcal{RC}_{Triples})$ 
    \Comment{Serialize the $\mathcal{RC}$ triples into text.}
    \State $GNN_{Ans} \gets \text{GNNPredict}(v_t,e_t)$ 
    \Comment{predict the top-K GNN answers for $v_t$ and $e_t$ }
    \State \hspace{-4ex} $P \gets \text{getPrompt}(Q,v_t,e_t,L,\mathcal{RC},GNN_{Ans},GLOW_v)$
    \Comment{generate the {\sysName}'s variation prompt}
    \State \Return $P$
\EndFunction
\end{algorithmic}
\label{algo:GLOW}
\end{algorithm}
\paragraph{Question Understanding \& Linking:}  
Given a question (e.g., \texttt{"What is the \textbf{species} of the protein \textbf{Q9LTJ2} from \textbf{BioKG}"?}), we extract the main entity node ($Q_n$), its type, the target relation ($Q_e$), and the KG name at algorithm~\ref {algo:GLOW} lines 2 and 3. The linking step at line 4 maps these elements to KG schema types using an LLM-based prompt (see Appendix A.2) over RDF metadata. For example, \texttt{species} maps to predicate URI \texttt{http://www.biokg.com/property/SPECIES}, denoted $e_t$, and \texttt{protein} maps to a schema node type URI. 
We resolve the question entity to node $v_t$ via a SPARQL query over name or label fields.

Unlike exact matching, this step uses semantic signals from attributes (name, label, description, URI) and schema-aware prompts to identify the node and relation types robustly.
For two-hop questions, the second edge is parsed and appended to the SPARQL query. \sysName{} generalizes to multi-hop reasoning through extended semantic parsing. 


\paragraph{KG Retrieval:} 
LLMs often lack domain-specific facts (e.g., in BioKG) to answer questions about $v_t$. However, neighboring KG nodes provide rich textual and structural context ($\mathcal{RC}$), which aids in inferring the missing label along $e_t$. The possible labels $L$ for edge $e_t$ are retrieved via SPARQL and provided to constrain generation at lines 5,6 and 7.
We retrieve 1-hop triples connected to $v_t$ via a SPARQL query generated at line 5 (\textit{Text-To-SPARQL}) that accounts for schema and namespace details and serialized into triples format at line 8.

Instead of retrieving top-K similar question entity examples like AskGNN~\cite{Ask-GNN}, which may be biased toward dominant GNN classes, we train a GNN NC model per question pattern (e.g., Protein$\rightarrow$Species). These GNNs are trained independently of the LLM and queried via an API at inference time to return top-$k$ candidate answers at line 9. See the GNN technical training/inference details in appendix \ref{GNN-details}. These predictions guide the LLM to correlate graph-derived candidates with the textual semantics in $\mathcal{RC}$, improving robustness. 
Unlike AskGNN, if the GNN underperforms due to poor structure, the LLM can still rely on $\mathcal{RC}$. This decouples model performance from GNN reliability.\shorten

\paragraph{Augmentation:}
Each prompt contains the question $Q$, node $v_t$, edge $e_t$, and possible answers $L$, optionally augmented with $\mathcal{RC}$ and GNN predictions at line 10. The prompt takes the form:
\begin{equation}
\hat{Y} = f(Q, v_t, e_t, L, \mathcal{RC}, \textit{GNN}_{Ans})
\end{equation}
Variant's Examples are provided in Appendix~\ref{app:prompts}.

\paragraph{Answer Generation:}
The LLM predicts a label for each node. Predictions are evaluated using an LLM-as-a-judge module described in \S\ref{subsec:llm-as-a-judge}.

\subsection{\sysName{} Pipeline Variants}
\label{sec:glowqa_variants}

We study how structured prompts influence QA using three \textit{In-Structure Learning} variants.

\paragraph{Graph-Context Prompt(\sysNameV{}-G):}
This variant adds the neighborhood $\mathcal{RC}$ of $v_t$, serialized into text via verbalization strategies~\cite{Triple_texualization} as shown in Figure \ref{fig:GLOWSys}.A. While this provides grounded semantic context, it may inflate prompt size if neighborhoods are large. See Figure~\ref{fig:GLOWSys} \textit{Text-To-SPARQL} query. AskGNN $\mathcal{RC}$ comprises the top-k similar question nodes as ICL examples, whereas \sysNameV{}-G $\mathcal{RC}$ includes a subgraph of attributes and neighboring nodes connected to the question node and serialized to offer contextual grounding to the LLM for effective reasoning.

\paragraph{A GNN-Guided Prompt (\sysNameV{}-N):}
As shown in Figure~\ref{fig:GLOWSys}.B, this variant injects top-$k$ GNN predictions for node $v_t$ as soft guidance. It provides structural signals without needing full KG serialization, but depends on GNN accuracy.
We train a GNN model for each question pattern using GraphSAINT. The training subgraphs are extracted using KGTOSA~\cite{KGTOSA}, excluding benchmark nodes. This improves scalability and yields diverse, task-specific subgraphs. At inference time, the GNN model returns top-$k$ candidates via API. See GNN details in the appendix.

\renewcommand{\arraystretch}{1.05}
\setlength{\tabcolsep}{3pt}
\begin{table*}[!t]
\centering
\caption{Summary of Open-World Question Template (OW-QT) across four KG-based features: (1) Knowledge Domain (KD): Domain-Specific (DS), Entertainment (E), or Generic (G); (2) Target Entity Type (ET); (3) Reasoning Hops (RH); and (4) Multiple Choice Count (MCC).}
\vspace{-0.2cm}
\label{tab:TaxonomyCompact}
\small
\begin{tabular}{@{}c c c c c c c c l@{}}
\toprule
\textbf{KG} & \textbf{KD} & \textbf{\#OW-QT} & \textbf{Target ET(s)} & \textbf{RH} & \textbf{\#Class} & \textbf{MCC} & \textbf{Label Types} \\
\midrule
YAGO4      & G  & 10 & Person \& Creative Work & 1–2 & 3–102  & 2–32+  & Nationality, Publisher, Occupation \\
BioKG      & DS & 6  & Drug \& Protein     & 1–2 & 2–29   & 2–32   & Kingdom, Class, SPECIES, R.Keyword \\
LinkedMDB  & E  & 5  & Film              & 1–2 & 7–39   & 4–32+  & Language, Producer, Genre \\
CrunchBase & DS & 4  & Investor          & 1   & 6–14   & 4–16   & Country, InvestRegion, Company \\
\bottomrule
\end{tabular}
\end{table*}

\paragraph{Hybrid Graph-GNN Prompt (\sysNameV{}-GN):}
This variant combines \sysNameV{}-G and \sysNameV{}-N by injecting both $\mathcal{RC}$ and GNN predictions. As shown in Figure~\ref{fig:GLOWSys}.C, it enables the LLM to reason jointly over semantic and structural cues. If GNN confidence is low, the LLM can still rely on the verbalized KG context.\shorten


\section{An Open-World Benchmark for KGQA} 
\label{sec:taxonomy}

We present {\Taxonomy}, a benchmark for multi-hop reasoning across diverse domains. It includes 25 open-world question templates based on four real-world KGs. 
Each template spans one of four dimensions: reasoning depth, knowledge domain, target entity type, and multiple-choice answer count.\shorten 
Table~\ref{tab:TaxonomyCompact} summarizes the templates by (see Appendix~\ref{app:Full_Bench} for details): 
1) \textit{Reasoning Hops (RH):} 1–2 steps from target to answer; 
2) \textit{Target Entity Type:} ranging from general (e.g., people, works) to domain-specific (e.g., drugs, proteins); 
3) \textit{Knowledge Domain (KD):} Generic (G, YAGO4~\cite{KG_Yago4}), Entertainment (E, LinkedMDB~\cite{LinkedMDB}), and Domain-Specific (DS, BioKG~\cite{biokg}, CrunchBase~\cite{KG_CrunchBase}); 
4) \textit{Multiple Choice Count (MCC):} 2–32+ candidates, one correct.\shorten 

While {\Taxonomy} builds on existing KGs, it introduces a new benchmark for OW-QA and multi-hop reasoning, which current KGQA datasets lack. All questions are designed with answers \emph{absent} from the KG, enabling realistic evaluation under incompleteness, and are grounded in real KGs but formulated for OW-QA.

\noindent \textbf{Task Formulation:}  
Each template defines a node classification task. For example, template \#1 classifies drugs by structure (Organic vs. Non-Organic) using BioKG. Given a target node $v_t$, the KG context is retrieved while excluding the gold answer (if present) from $\mathcal{RC}$. The model selects the correct answer from the candidate set.

\noindent \textbf{Answer Evaluation:}  
\label{subsec:llm-as-a-judge}
Evaluating LLM predictions is challenging due to linguistic variation, making exact matching insufficient. We adopt the LLM-as-a-Judge framework~\cite{LLMAsJudge}, where an auxiliary LLM compares outputs to gold answers in two modes:
1) \textit{Exact Match (EM):} Identical or semantically equivalent terms (e.g., \textit{Actor} vs. \textit{Film Star});  
2) \textit{Hierarchical Match (HM):} Synonyms or subtypes (e.g., \textit{Athlete} and \textit{Player}).  
We use GPT-4o-mini as the evaluation judge~\cite{JudgeBench}.

\section{Experiments}
\label{sec:evaluation}
\begin{table*}[]
\caption{
\textbf{Exact Match Accuracy (\%) of {\sysName} vs. baselines across four OWA-QA datasets.} {\sysNameV}-GN consistently outperforms AskGNN, GCR, GoG, and LLM-only setups across all LLMs and datasets, showing strong generalization and effective use of both textual and structural signals, even with smaller LLMs.
}
\centering
\label{tab:exp_avg_res_qwen3-8b}
\small
\resizebox{0.995\textwidth}{!}{%
\begin{tabular}{clcc|cccc|ccc}
\toprule
 & \textbf{LLM-Model} & \textbf{Dataset} & \textbf{RH} & \textbf{LLM-Only} & \textbf{AskGNN} &\textbf{GCR}&\textbf{GoG}& \textbf{GLOW-G} & \textbf{GLOW-N} & \textbf{\makecell{GLOW-GN}} \\ 
 \hline
 \multirow{15}{*}{\rotatebox{90}{\textbf{Open-Weight LLMs}}} & \multirow{5}{*}{\textbf{\makecell{Qwen3-8B \\ ~\cite{qwen3-8b}}}} & Arxiv2023 & 1 & 31  & 62 &4&53&64 & 66 & \textbf{81} \\
 & & ogbn-arxiv & 1 & 12  & 70 &14&57& 71 & 60 & \textbf{77} \\
 & & ogbn-product & 1 & 34  & 29 &5&44& 45 & 51 & \textbf{57} \\
 & & \Taxonomy & 1 & 30  & 79 &34&48& 60 & 80 & 83\\
 & & \Taxonomy & 2 & 19  & 34 &15&29& \textbf{44} & 37 & 42 \\
 \cline{2-11} 
 & \multirow{5}{*}{\textbf{\makecell{DeepSeek-R1-\\Distill-Qwen-7B \\ ~\cite{deepseek-r1-7b}}}} & Arxiv2023 & 1 & 24  & 48 &7&28& 55 & 59 & 67 \\
 & & ogbn-arxiv & 1 & 13  & 50 &15&35&30 & 61 & 63 \\
 && ogbn-product & 1 & 27  & 19 &12&44&38 & 43 & 51 \\
 & & \Taxonomy & 1 & 21  & 75 &36&20& 58 & 78 & \textbf{81} \\
 & & \Taxonomy & 2 & 12 & 31 &16&13&27 & 32 & \textbf{39} \\
 \cline{2-11} 
 & \multirow{5}{*}{\textbf{\makecell{Granite-3.3-8B-\\Instruct\\ ~\cite{Granite-8b}}}} & Arxiv2023 & 1 & 14&   63 &14&53& 32 & 63 & 77 \\
 & & ogbn-arxiv & 1 & 5  & 66 &21& 56&31 & 55 & \textbf{73} \\
 & & ogbn-product & 1 & 22  & 30 &13&47 &45 & 52 & 54\\
 & & \Taxonomy & 1 & 24 & 79 & 47&42&58 & 80 & \textbf{82} \\
 & & \Taxonomy & 2 & 14  & 35 & 19&27&39 & 32 & \textbf{42} \\
\midrule
 \multirow{10}{*}{\rotatebox{90}{\textbf{Commercial LLMs}}} & \multirow{5}{*}{\textbf{\makecell{GPT-4o-mini\\ ~\cite{gpt40mini}}}} & Arxiv2023 & 1 & 35  & N/A &N/A &41& 59 & 48 & 62\\
 & & ogbn-arxiv & 1 & 50 & N/A &N/A& 58&63 & 55 & \textbf{67} \\
 & & ogbn-product & 1 & 37  & N/A & N/A&48& 43 & 51 & \textbf{59} \\
 & & \Taxonomy & 1 & 29  & N/A &N/A& 57& 68 & 78 & \textbf{82} \\
 & & \Taxonomy & 2 & 28  & N/A &N/A& 31& \textbf{53} & 35 & 49 \\
 \cline{2-11} 
 & \multirow{5}{*}{\textbf{\makecell{DeepSeek-V3 \\ ~\cite{deepseek-llm}}}} & Arxiv2023 & 1 & 17  & N/A &N/A &44& 57 & 61 & 73 \\
 & & ogbn-arxiv & 1 & 58  & N/A &N/A& 61& 62 & 63 & \textbf{65} \\
 & & ogbn-product & 1 & 36  & N/A& N/A &51& 45 & 50 & \textbf{57} \\
 & & \Taxonomy & 1 & 34  & N/A &N/A&60& 69 & 78 & \textbf{84} \\
 & & \Taxonomy & 2 & 30  & N/A &N/A&47& \textbf{54} & 34 & 53 \\
\bottomrule
 \end{tabular}}
 \end{table*}

\subsection{Experimental Setup}
\nsstitle{Datasets and Metrics:}  
In addition to our {\Taxonomy}, we also evaluate \sysName{} on three existing datasets ~\cite{Ask-GNN}; arxiv2023, ogbn-arxiv, and ogbn-products. These datasets are originally node classification datasets, but AskGNN~\cite{Ask-GNN} adopted it for OW-QA, where each dataset is converted into a QA dataset using predefined question templates. In all experiments, we report the average across two runs. \shorten

\nsstitle{Baselines:}  
We compare \sysName{} against recent methods in 3 categories: \textit{LLM-Only}, \textit{Open-world KGQA}, and \textit{closed-world KGQA}. 

\textbf{LLM:} we use commercial models like GPT-4o-Mini~\cite{gpt40mini} and DeepSeek-V3~\cite{deepseekai2025deepseekv3technicalreport}, open-weight models including Qwen3-8B~\cite{qwen3-8b}, DeepSeek-R1-Distill-Qwen-7B~\cite{deepseek-r1-7b}, and IBM Granite 3.3-8B-Instruct~\cite{Granite-8b}.

\textbf{Open-world KGQA}: \textit{AskGNN}~\cite{Ask-GNN} is an approach that integrates LLMs and GNNs for QA over homogeneous graph datasets. We adapt \textit{AskGNN} for the OW-QA setting over KGs and evaluate its performance on more challenging benchmark datasets. Recently \textit{GoG} \cite{Generate-on-Graph} has been developed as a KGQA system designed for incomplete KGs where the answer path is intentionally partially removed.  
\\

\textbf{Closed-world KGQA:} \textit{GCR}~\cite{GCR} is a method designed for scalable KGQA over large KGs. It outperforms \textit{RoG}~\cite{RoG} via fine-tuning LLMs to extract answers from retrieved answer paths.\shorten

\nsstitle{Evaluation Setup:}  
GNN training was performed on an Ubuntu VM with dual 32-core Intel Xeon 2.4GHz CPU, 250GB RAM, and V100D-8C 16G vGPU. The GNN models were trained for the node classification tasks using GraphSAINT~\cite{GraphSAINT} and ShaDowGNN \cite{ShaDowGNN} with the task-oriented sampling method in ~\cite{KGTOSA}. KGs were hosted on Virtuoso 07.20.32 with default settings. The GCR 8B models are fine-tuned using Colab A100 GPUs with 40G of VRAM.   \shorten


\subsection{Experimental Results}


\textbf{Benchmark Results:}  
As shown in Table~\ref{tab:exp_avg_res_qwen3-8b}, {\sysNameV} consistently outperforms all baselines across LLMs and datasets, showing strong generalization and effective use of textual and structural cues. Unlike AskGNN, which depends on GNN-based ICL examples, {\sysNameV} retrieves the question’s entity context and combines GNN predictions with neighborhood text for better performance.
For example, using Qwen3-8B, {\sysNameV}-GN outperforms AskGNN by 18\% on average across the AskGNN datasets, and by 4\% and 8\% on 1-hop and 2-hop {\Taxonomy} questions, respectively. It also surpasses GCR by (64\%, 49\%, and 27\%) and GoG by (20\%, 35\%, and 13\%) . Similar trends hold across other open-weight (e.g., DeepSeek-R1-Distill-Qwen-7B, Granite-3.3-8B-Instruct) and commercial (e.g., GPT-4o-mini, DeepSeek-V3) LLMs.\shorten

AskGNN's performance varies and declines when either the GNN or LLM underperforms, showing its reliance on both components. GCR depends entirely on the presence of correct answer paths; when missing, it generates incorrect paths and answers. GoG relies entirely on generating missing paths, which are not constrained by the underlying KG schema, often leading to the creation of misleading or semantically invalid answer paths. In contrast, {\sysNameV} remains robust even with smaller LLMs, thanks to effective retrieval and contextualization. More detailed results per template and LLM are shown in Appendix A.5.
For generic questions like Person → Nationality, textual attributes (e.g., name, residence, education) often suffice for LLMs to infer the answer. But for domain-specific tasks (e.g., Protein → Family or Protein → Species), textual cues are limited. Structurally similar nodes help the GNN yield better predictions.\shorten 

\begin{figure}[!t]
  \centering
\includegraphics[width=0.995\linewidth]{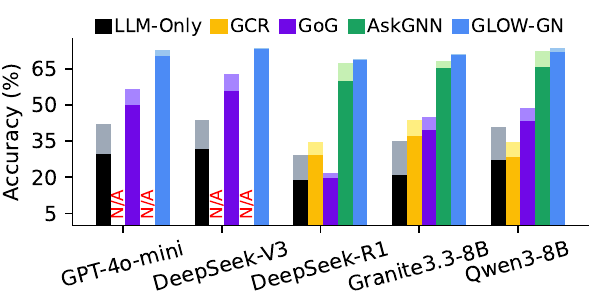}
  \caption{
  Average Exact and Hierarchical Match Accuracy (\%) across datasets using different methods and LLMs. Dark bars show Exact Match; lighter segments show added Hierarchical Match gains. GCR and AskGNN are inapplicable to commercial LLMs.\shorten
 }
  \label{fig:GLOW_Exact_Vs_Her}
   \vspace{-1em}
\end{figure} 

\begin{figure*}[!t]
  \centering
  \includegraphics[width=0.99\linewidth]{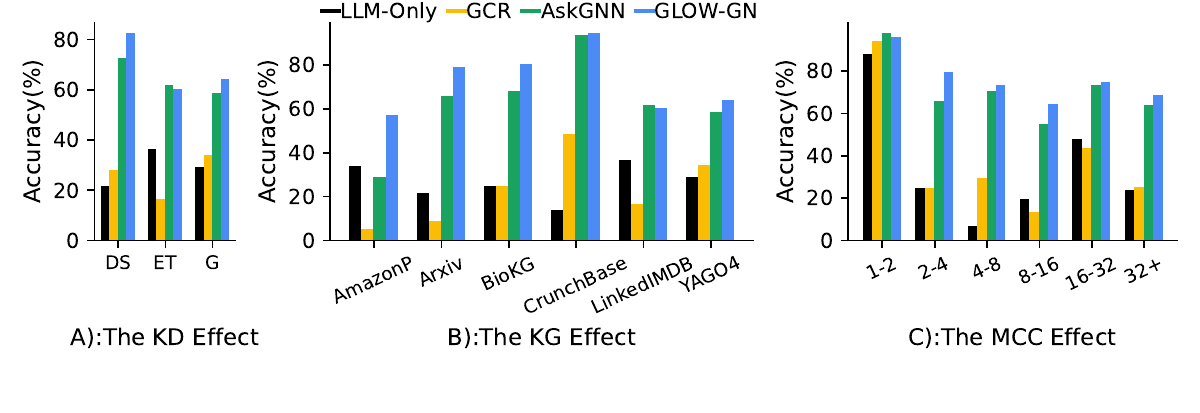}
  \vspace{-1.5em}
  \caption{Effect of {\Taxonomy} characteristics on the \sysName{} answer accuracy(\%) with Qwen3-8B. The effects are grounded by A) Knowledge Domain (KD), B) Knowledge Graph (KG) , and C) Multiple Choice Count (MCC). \shorten }
  \label{fig:GLOWBench-Characteristics}
\end{figure*}

Combining both inputs in {\sysName}-GN consistently boosts accuracy—up to 26 points on ogbn-arxiv and ogbn-products, and 6 points on {\Taxonomy} templates like creative-work→country. {\sysName}-G performs best on 2-hop {\Taxonomy} with GPT-4o-mini and DeepSeek-V3 due to weaker GNN results on dense subgraphs. For example, in creative-work→genre, the GNN reached only 22\% vs. 33\% by {\sysName}-G. In contrast, for drug→class, GNN accuracy was higher (68\% vs. 45\%).\shorten

\nsstitle{Exact vs. Hierarchical Match Accuracy:} 
LLMs paraphrase answers or return semantically related concepts rather than producing exact matches. For instance, the occupation "Singer" may be returned in place of "Artist", its superclass—potentially acceptable in some contexts. Figure~\ref{fig:GLOW_Exact_Vs_Her} analyzes this phenomenon by comparing Hierarchical-Match and Exact-Match accuracies, where GPT-4o-mini is used as a judge for the Hierarchical-Match. \shorten

The $LLM_{Only}$ pipeline, which relies purely on pretraining without structural grounding, is particularly prone to generating such approximate answers. On average, its Hierarchical-Match accuracy exceeds Exact-Match accuracy by 12.5\%, reflecting this tendency. This gap narrows to 3.3\% with AskGNN, which supplements the LLM with contextual graph signals but still lacks fine-grained control over output specificity.
In contrast, {\sysNameV}-GN demonstrates minimal reliance on hierarchical leeway—showing only a 1.1\% gain—indicating its robustness in steering the LLM toward precise answers. The integration of textual and structural semantics helps the LLM disambiguate fine-grained targets, improving accuracy and consistency across datasets and LLM architectures.

\subsection{Effect of Domain, Graph, and Question}

\nsstitle{Knowledge Domain (KD):}
We analyze model performance across question domains using {\Taxonomy}. As shown in Figure~\ref{fig:GLOWBench-Characteristics}.A, {\sysNameV}-GN consistently outperforms all baselines. Gains are most notable in domain-specific (DS) areas such as pharmaceuticals and proteins, where {\sysNameV}-GN significantly exceeds models like Qwen3-8B. These results reflect LLMs’ difficulty in handling fine-grained, specialized knowledge. By integrating textual and structural signals, {\sysNameV}-GN enables stronger generalization. Even in general domains like Generic (G) and Entertainment (ET), where entities are likely seen during pretraining, {\sysNameV}-GN maintains an edge—often rivaling AskGNN without requiring additional fine-tuning.

\nsstitle{KG Structure:}
We assess the effect of KG structure by comparing performance across KGs used in Amazon-Product, Arxiv, and four {\Taxonomy} KGs (BioKG, CrunchBase, LinkedIMDB, YAGO4). Figure~\ref{fig:GLOWBench-Characteristics}.B shows that {\sysNameV}-GN outperforms baselines on all domain-specific KGs and performs well on YAGO4, a generic KG. The gap narrows only on LinkedIMDB, likely due to high entity overlap with LLM training corpora.

\begin{figure}[!t]
  \centering
    \vspace{-0.5em}
    \includegraphics[width=0.995\linewidth]{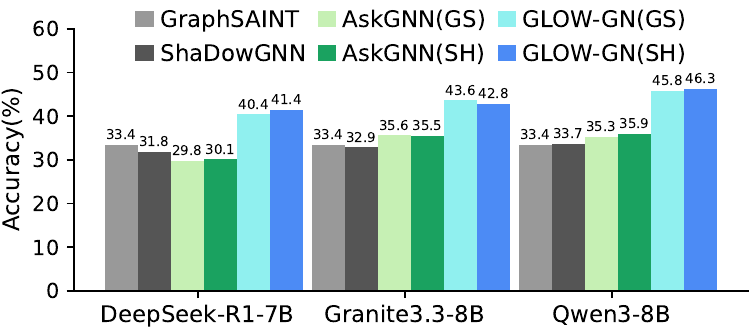}
  \vspace{-1em}
  \caption{The GNN error propagation using GraphSAINT(GS) and ShaDowGNN(SH) models on questions with accuracy score  $<$ 50\%. \sysName{}  boosts accuracy by up to 12\% over weak GNNs using textual cues. AskGNN closely follows GNN performance.}  
  \label{fig:GLOW_GNN_dependency}
  \vspace{-1em}
\end{figure}

\nsstitle{Answer Choices (MCQ Format):}
We evaluate robustness under varying MCQ settings. Prior work~\cite{LLM-MCQ} showed LLM accuracy drops as the number of choices increases, and Figure~\ref{fig:GLOWBench-Characteristics}.C confirms this. Still, {\sysNameV}-GN retains its edge, especially beyond two choices, underscoring the value of graph-based disambiguation under higher decision complexity.

\vspace{-2ex}

\subsection{GNN Models and Answer Selection}
\nsstitle{GNN Error Propagation Analysis:}
AskGNN’s performance depends heavily on GNN quality. When the GNN is weak, AskGNN provides little to no gain and may even underperform the GNN itself. Figure~\ref{fig:GLOW_GNN_dependency} shows overall QA accuracy on OWA questions on which their corresponding GraphSAINT and ShaDowGNN models (used by AskGNN) scored below 50\%. Across several LLMs, AskGNN typically matches or lags behind the GNN baseline. 
In contrast, {\sysNameV}-GN combines textual semantics and retrieval-based reasoning to outperform weak GNNs, with gains up to 12\%. For example, on Amazon-product with Qwen3-8B, GraphSAINT scored 45\%, AskGNN dropped to 29\%, while {\sysNameV}-GN reached 51\%. On Protein-Keyword, AskGNN achieved 35\%, GraphSAINT 42\%, and {\sysNameV}-GN improved to 57\%.\shorten

\begin{figure}[!t]
  \centering
  \vspace{-0.7em}
  \includegraphics[trim=0cm 0cm 0cm 1cm, clip=true, width=0.99\linewidth]{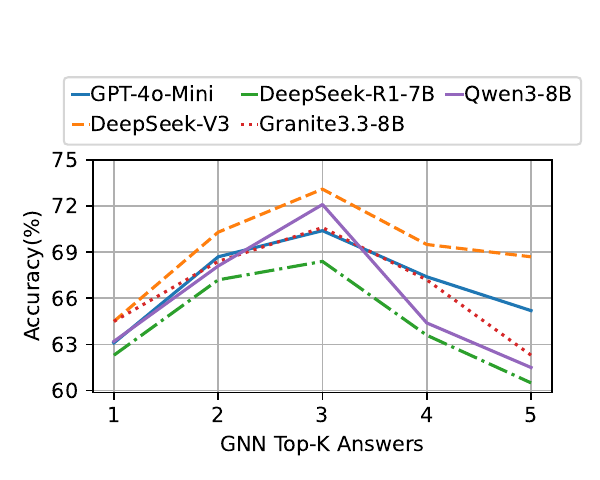}
  \vspace{-1.5em}
  \caption{Impact of GNN top-K answer count on {\sysName}-GN’s average accuracy (\%) across different LLMs. {\sysName}-GN achieves the best performance at top-K=3, while higher values tend to mislead the LLMs.}
  \label{fig:GLOW_GNN_TopK}
\end{figure}

\vspace{0.5em}
\nsstitle{Varying GNN Top-K Answers:} 
Increasing top-$K$ GNN answers confuses the LLM, lowering accuracy. Figure~\ref{fig:GLOW_GNN_TopK} shows {\sysName}-GN performs best at $K=3$ across five LLMs, while $K=4$ or $5$ reduces performance. Large closed-weight LLMs are less sensitive to higher $K$ values.

\subsection{System Efficiency Analysis}
\nsstitle{Training Time:} AskGNN’s joint GNN–LLM training grows with graph size, structural complexity (e.g., node/edge types), and LLM scale, averaging 23.4 hours across tasks. GCR, fine-tuned on WebQSP, takes about 6 GPU hours for 500 epochs. In contrast, {\sysNameV}-N and {\sysNameV}-GN decouple GNN and LLM training, requiring just 1.7 hours for GNN training (Table~\ref{tbl:Time-Token}). This modular design enhances scalability and avoids dependence on LLM size, easing adaptation to new domains.

\vspace{1em}
\nsstitle{Token Count:} We analyze average token consumption per question across all pipelines. {\sysNameV}-N incurs the lowest token usage at 0.43K tokens per prompt, as its retrieved context ($\mathcal{RC}$) consists solely of GNN-predicted answers and candidate labels, mirroring AskGNN's structure but with reduced verbosity. 
GCR incurs the highest token cost due to XML-formatted triples. GoG incurs the highest token cost due to performing an agentic chain-of-thought rounds for missing triple generation. AskGNN averages 0.79K tokens, driven by 20 ICL examples; fewer examples significantly reduce performance. {\sysNameV}-G and {\sysNameV}-GN offer a balanced cost, with prompt length shaped by KG density and the number of neighbors connected to the question node.\shorten

\setlength{\tabcolsep}{5pt} 
\setlength{\arrayrulewidth}{0.1mm}
\begin{table}[!t]
\centering
\caption{The average training time in Hours, Answer tokens count per question in (K-Tokens) and answer time per question in Seconds using Qwen3-8B LLM.}
\label{tbl:Time-Token}
\small
\begin{tabular}{l|ccc}
\hline
&\textbf{\makecell{Training\\Time (H)}} &\textbf{\makecell{Tokens\\Count (K)}} &\textbf{\makecell{Answer\\Time (Sec)}} \\
\hline
\textbf{AskGNN} &23.4 &0.79 &11.2 \\
\textbf{GCR} &6&0.84&13.4 \\
\textbf{GoG} &N/A&1.5&15.7 \\
\hline
\textbf{{\sysName}-G} &N/A &0.78 &12.6 \\
\textbf{{\sysName}-N} &\textbf{1.7} &\textbf{0.43} &\textbf{8.2} \\
\textbf{{\sysName}-GN} &\textbf{1.7} &0.65 &11.5 \\
\hline
\end{tabular}
\end{table}
\vspace{1em}
\nsstitle{Question Answering Time:} 
In Table \ref{tbl:Time-Token}, {\sysNameV}-N is the fastest, averaging 8.2s per question due to its compact prompt. GCR is slower, as it relies on retrieved paths—often missing in OW-QA—causing hallucinations and longer processing. GoG is slowest, performing agentic chain-of-thought rounds for missing triple generation. AskGNN follows at 11.2s, hindered by reasoning over ICL examples. {\sysNameV}-GN shows moderate latency, combining GNN outputs with subgraph context. All experiments use vGPUs, not A100/H100; faster hardware would likely reduce runtime.

\vspace{1em}
\begin{figure}[!h]
  \centering
    \includegraphics[width=0.995\linewidth]{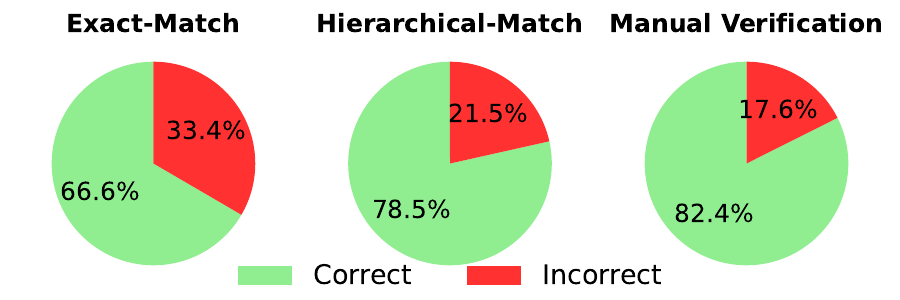}
  \caption{Human evaluation results versus Exact-match and Hierarchical-match. Hierarchical matching closely aligns with human evaluation.}  
  \label{fig:GLOW_Manual_verf}
\end{figure}
\nsstitle{Human Based Evaluation:}
To evaluate our LLM-as-a-Judge prompts, we manually validated the Qwen3:8B {\sysNameV}-GN answers across the {\Taxonomy} question patterns, using 5 randomly selected questions per pattern. As shown in Figure \ref{fig:GLOW_Manual_verf}, The Exact-match accuracy reached 66.6\%, Hierarchical match 78.5\%, and Manual verification is 82.4\%. Notably, Hierarchical matching recovered 95.2\% of manually verified correct answers, demonstrating the robustness of our prompts. Some responses were marked as non-matches hierarchically despite being valid (e.g., (CEO, Co-Founder), (20th Century Studios, Paramount Pictures), (Catalan, Spanish)), while a few cases were incorrectly flagged as non-exact matches (e.g., (American English, English)).\shorten
\vspace{1em}

\section{Conclusion}\label{sec:cls}

This paper introduces {\sysName}, a system for open-world QA on KGs that integrates LLMs with GNNs. {\sysName} uses GNN-predicted candidates and relevant subgraphs as structured context to enhance multi-hop reasoning over incomplete KGs. By combining structural and textual semantics, {\sysName} overcomes key limitations of closed-world and retrieval-based KGQA. It consistently achieves strong performance across question types, domains, and LLMs, even when component quality varies. On standard open-world benchmarks and our new {\Taxonomy} dataset, {\sysName} shows significant improvements in exact and semantic accuracy. These results highlight the need for hybrid approaches tailored to open-world settings and confirm {\sysName}’s robustness and generalizability.


\section{Limitations}
This work has three main limitations: First, data quality issues, such as sparse KGs with limited node and edge descriptions, impair both textual and structural semantics, reducing performance, especially for large LLMs. Second, our approach depends on high-performing GNN models for effectiveness. Third, all questions are currently framed as node classification tasks, though some may be better suited to link prediction, requiring prior evaluation and task-specific formulation.

\bibliography{paper}
\newpage
\appendix
\onecolumn
\section{Appendix}
\label{sec:appendix}

\subsection{{\sysName} Prompt Examples}
\label{app:prompts}
 
\textbf{User Prompt}:
\begin{boxE}
{
\label{prompt_LLLMOnly}
\small
Predict the chemical kingdom for the drug Yohimbine from the BioKG knowledge graph.\\
Answer:
}
\end{boxE}
\textbf{Question Understanding Prompt}:
\begin{boxE}
{
\label{prompt_QUnderstand}
\small
<system>: You are an an expert Entity-Extraction NLP system.\\
<user>: Given the following question, identify 1-The question main entity type, 2- the main entity, 3- the prediction label and \\4- the KG name.\\
Question: \{\}\\
Answer:\\
1-Question main entity:  2-Main Entity:\\   
3-Prediction label: 4-KG name:
}
\end{boxE}
\textbf{Entity/Relation Linking Prompt}:
\begin{boxE}
{
\label{prompt_ERL}
\small
<system>: You are an an expert knowledge graph entity-relation Linking NLP system.\\
<user>: Given the following KG Schema in the basic graph pattern CSV format, one predicate per line:
node type, relation, node type. \\
-----------\\
KG Schema:\\\{\}\\
------------\\
1- What is the node type in the schema that corresponds to \{main entity\}? Return only the name.\\
2- Choose from the schema the BGP (node type, relation, node type) that describes the  \{main entity type\} value \{main entity\}. Return only the BGP.\\
3- Choose from the schema the BGP 
(node type, relation, node type) that describe the  \{main entity type \} \{prediction label\}. \\
Return only the BGP.\\
------------\\
Answer:\\
1-\\
2-\\
3-
}
\end{boxE}
\newpage
\textbf{Question To SPARQL Prompt}:
\begin{boxE}
{
\label{prompt_QSPARQL}
\small
<system>: You are an an expert text-To-SPARQL translation system.\\
<user>: Given the following KG Schema in the basic graph pattern CSV format, one predicate per line:
node type, relation, node type. \\
-----------\\
KG Schema:\\\{\}\\
------------\\
Write a SPARQL query that selects the \{main entity type\} that satisfy the following BGPs.\\
1-  \{question node type\} ,[label/name/titile] ,\{question node\}\\
2-  \{Other BGPs\}\\
graph prefix: \{KG Prefix\}\\
--------------------\\
Answer: SPARQL Query\\
Do not return any explanation or reasoning details.
}
\end{boxE}

\textbf{SPARQL Query Example}:
\begin{boxE}
{
\label{prompt_SPARQL_example}
\small
PREFIX biokg: <http://www.biokg.com/>\\
SELECT ?drug as ?vt  ?kingdom as ?vl\\
WHERE \{\\
  VALUES ?name \{ "Yohimbine" \}\\
  ?drug biokg:NAME ?name .\\
  ?drug biokg:KINGDOM ?kingdom .\}
}
\end{boxE}

\textbf{Basic Prompt}:
\begin{boxE}
{
\label{prompt_L}
\small
<system>: You are an expert open world question answering system.\\
<user>: What is the \textbf{$\{$Prediction Label$\}$} of the \textbf{$\{$question entity type$\}$} \textbf{$\{$question entity$\}$} from \textbf{$\{$KG details$\}$} knowledge graph.  
\\- Do not return any context or analysis.
\\- Help: The possible list of \textbf{$\{$Prediction Label Type$\}$}s are: [\textbf{$\{$ Labels List$\}$}]\\
Answer
}
\end{boxE}

\vspace{-1em}

To generate an instance of this template, replace the question node type, KG, and prediction label with values from one of the queries in table \ref{tab:Taxonomy}. 
\textbf{An example prompt for OWA-Q \#1}:

\begin{boxE}
{\small
<system>: You are an expert open world question answering system.\\
<user>: What is the \textbf{Kingdom} of the  \textbf{Drug} Yohimbine from the \textbf{BioKG , a Biomedical} knowledge graph.  
\\- Do not return any context or analysis.
\\- Help: The list of \textbf{Kingdoms} are: [Organic,Non-Organic]    \\
Answer:
}\end{boxE}

\textbf{{\sysNameV}-G Prompt}:
\begin{boxE}
{ \small
<system>: You are an expert open world question answering system.\\
<user>: What is the \textbf{Kingdom} of the  \textbf{Drug} Yohimbine from the \textbf{BioKG , a Biomedical} knowledge graph.  
\\- Do not return any context or analysis.
\\- Help: The possible list of \textbf{Kingdoms} are: [\textbf{Organic,Non-Organic}]
\\- The \textbf{Drug} associated  triples.
 [("Yohimbine","DDI", "DB13677"), ..]\\
Answer:
}
\end{boxE}
\newpage
\textbf{{\sysNameV}-N Prompt}:
\begin{boxE}
{\small
<system>: You are an expert open world question answering system.\\
<user>: What is the \textbf{Kingdom} of the  \textbf{Drug} Yohimbine from the \textbf{BioKG , a Biomedical} knowledge graph. 
\\- Do not return any context or analysis.
\\- Help: The possible list of \textbf{Kingdoms} are: [\textbf{Organic,Non-Organic}]
\\-  Verify the following list of GNN  Answers: [ Organic,Non-Organic, ..]\\
Answer:
}
\end{boxE}

\textbf{{\sysNameV}-GN Prompt}:
\begin{boxE}
{\small
<system>: You are an expert open world question answering system.\\
<user>: What is the \textbf{Kingdom} of the  \textbf{Drug} Yohimbine from the \textbf{BioKG , a Biomedical} knowledge graph. 
\\- Do not return any context or analysis.
\\- Help: The possible list of \textbf{Kingdoms} are : [\textbf{Organic,Non-Organic}]
\\- Verify the following  GNN  Answer: [ Organic]
\\- The \textbf{Drug} associated  triples.
 [("Yohimbine","DDI", "DB13677"), ..]\\
Answer:
}
\end{boxE}
\subsection{LLM as-a-Judge Prompt}
\begin{boxE}
{
\label{prompt_llm_as_judge}
\small
<system>: You are an expert LLM-as-a-Judge system.\\
<user>: Given the following list of predicted and true pairs of values.\\ 
-Rank the predicted value against the true value using two metrics.\\
1- Exact Match Rule: you compare the two strings after normalization and remove any special characters. report 1 if both values are literally and semantically equal and 0 otherwise.\\
2- Hierarchical/Categorical Match Rule: report 1 if the predicted value is under a subcategory or hierarchically belongs to the true value or is a synonym and 0 otherwise.\\
- Example:\\
List of pairs: [[music, art], [painter, artist],[ football player, soccer player], [ lawyer, judge], [lawyer, player]]\\
Answer: [[0,1],[0,1],[1,1],[0,1],[0,0]]\\
- Question:\\
-List of pairs: \textbf{$\{$ListOfPairs$\}$}\\
-Note: refine each pair and return Answer for exactly \textbf{$\{$length(ListOfPairs)$\}$} pairs without explanation.\\
-Finally: make sure you return only \textbf{$\{$length(ListOfPairs)$\}$} pair of answers.\\
Answer:
}
\end{boxE}
\newpage
\subsection{Full Details of Our {\Taxonomy} }
\label{app:Full_Bench}
\renewcommand{\arraystretch}{1.15}
\setlength{\tabcolsep}{4pt}

\begin{table*}[!h]
\centering
\caption{Our {\Taxonomy} benchmark characterizes each Open-World Question Template (OW-QT) across four KG-based features: (1) Knowledge Domain (KD): Domain-Specific (DS), Entertainment (E), or Generic (G); (2) Target Entity Type (ET); (3) Reasoning Hops (RH); and (4) Multiple Choice Count (MCC).}
\label{tab:Taxonomy}
\small
\begin{tabular}{@{}c c c l c c c l l@{}}
\toprule
\textbf{OW-QT} & \textbf{KG} & \textbf{KD} & \textbf{Target ET} & \textbf{RH} & \textbf{\#Class} & \textbf{MCC} & \textbf{Label Type} & \textbf{Reasoning Path} \\
\midrule

1  & \multirow{6}{*}{\rotatebox{90}{BioKG}}    & DS & Drug    & 1 & 2  & 2–4   & Kingdom     & Kingdom \\
2  &                                          & DS & Drug    & 1 & 13 & 8–16  & Superclass  & Superclass \\
3  &                                          & DS & Drug    & 2 & 29 & 16–32 & Class        & RelatedPubMed $\rightarrow$ Class \\
4  &                                          & DS & Protein & 1 & 18 & 16–32 & SPECIES      & SPECIES \\
5  &                                          & DS & Protein & 2 & 15 & 8–16  & R.Keyword    & RelatedPubMed $\rightarrow$ R.Keyword \\
6  &                                          & DS & Protein & 1 & 4  & 2–4   & Family       & Family \\
\midrule

7  & \multirow{4}{*}{\rotatebox{90}{CrunchBase}} & DS & Investor & 1 & 12 & 8–16 & Country       & Country \\
8  &                                              & DS & Investor & 1 & 14 & 8–16 & InvestRegion  & InvestRegion \\
9  &                                              & DS & Investor & 1 & 6  & 4–8  & PositionTitle & PositionTitle \\
10 &                                              & DS & Investor & 1 & 11 & 8–16 & Company       & Company \\
\midrule

11 & \multirow{5}{*}{\rotatebox{90}{LinkedMDB}} & E & Film & 1 & 15 & 8–16  & Language  & Language \\
12 &                                            & E & Film & 1 & 27 & 16–32 & Country   & Country \\
13 &                                            & E & Film & 2 & 39 & 32+   & Producer  & Sequel $\rightarrow$ Producer \\
14 &                                            & E & Film & 2 & 7  & 4–8   & Genre     & Sequel $\rightarrow$ Genre \\
15 &                                            & E & Film & 1 & 28 & 16–32 & Subject   & Subject \\
\midrule

16 & \multirow{10}{*}{\rotatebox{90}{YAGO4}} & G & CreativeW & 2 & 13  & 8–16  & ProdComp     & byArtist $\rightarrow$ ProdComp \\
17 &                                         & G & CreativeW & 2 & 14  & 8–16  & Publisher    & isBasedOn $\rightarrow$ Publisher \\
18 &                                         & G & CreativeW & 1 & 25  & 16–32 & PublishLang  & Author \\
19 &                                         & G & CreativeW & 2 & 11  & 8–16  & Genre        & ProdComp $\rightarrow$ Genre \\
20 &                                         & G & CreativeW & 1 & 6   & 4–8   & Country      & Country \\
21 &                                         & G & Person    & 1 & 3   & 2–4   & GivenAward   & GivenAward \\
22 &                                         & G & Person    & 1 & 91  & 32+   & Nationality  & Nationality \\
23 &                                         & G & Person    & 1 & 4   & 4–8   & GraduateOfOrg & GraduateOfOrg \\
24 &                                         & G & Person    & 1 & 102 & 32+   & Occupation   & Occupation \\
25 &                                         & G & Person    & 1 & 8   & 4–8   & SpokenLang   & SpokenLang \\
\bottomrule
\end{tabular}
\end{table*}

\subsection{The GNN Training/Inference Technical Details.}
\label{GNN-details}
\begin{itemize}
\item For each question pattern, we train a node classification GNN using GraphSAINT with RGCN as GNN convolutional layer to support heterogeneous KG subgraphs. 
The initial node embeddings are iteratively aggregated with embeddings received from neighboring nodes connected to a specific relation until the embeddings of all nodes converge.  The final embedding of a $v_t$, our main question entity, is obtained through two aggregations:  an outer aggregation over each relation type and an inner aggregation over neighbouring nodes $\mathcal{RC}$ of a specific relation and defined by RGCN~\cite{RGCN} as follows:
\begin{equation}
\label{eq_rgcn}
h_i^{(l+1)}= \sigma(\sum_{r \in R}\sum_{j \in N_i^r}\frac{1}{c_{i,r}}W_r^{(l)}h_j^{(l)}+W_0^{(l)}h_i^{(l)}) 
\end{equation}

where $l$ is an RGCN layer, $h^{(l+1)}_j$ is the hidden embedding of node $j$ at layer $l+1$, $\sigma$ is element-wise activation function, $N_i^r$ denotes the set of neighbour indices of node $i$ under relation $r \in \mathcal{R}$, $c_{i,r}$ is a normalization constant that can either be learned or chosen in advance (such as $c_{i,r}=|N_i^r|$), $W^{(l)}_r$ is the weight matrix for relation $r$ at layer $l$, and $W^l_0$ is the initial weight matrix at layer $l$.\shorten

\item To construct the training set, we exclude the {\Taxonomy} question entity nodes from the training set and extract 1-hop in and out neighbor nodes subgraphs using the KGTOSA sampler \cite{KGTOSA}. This ensures scalability to large KGs and yields task-relevant and diverse subgraphs for effective training.
\item GraphSAINT Training HyperParameters: We set the input dimension (D = 128), hidden channels dimension=64, and number of layers (L = 2) of our GNN module with dropout rate 0.5 applied to each layer. We train the model with the Adam optimizer,
learning rate from {5e-4, 5e-3, 1e-3, 2e-3}.
\item  Subgraph Sampling Hyperparameters:  batch-size=20000, walk-length=2, num-steps=10. \shorten
\item At inference time, the node classification model M is resolved using the KG name, the question's main entity type, and the target edge. The model M predicts the top-k answer classes based on the Log-Likelihood for each class and feeds them to the LLM prompt.
The trained GNN model is hosted via an inference API. At runtime, this API returns the top-K predicted answer nodes, which are then passed to the {\sysName} pipeline as supportive evidence for the LLM’s reasoning.
\end{itemize}

\subsection{The OWA-QA Detailed Results:}
\definecolor{rowgray}{gray}{0.95}
\begin{table}[!htp]\centering
\rowcolors{2}{rowgray}{white}
  \caption{The detailed accuracy (\%) of {\sysName} compared to baseline models on existing question answering datasets from \cite{Ask-GNN} and our developed benchmark {\Taxonomy}. All results below use Qwen3-8B as the underlying LLM. Best results are marked in \textbf{bold}. }
\scriptsize
\begin{tabular}{ccc|cccc|ccccc}\toprule
\label{tab:exp_res_qwen3-8b}
LLM-Model &Dataset &RH &AskGNN &GraphSAINT &LLM-Only&GCR &\textbf{GLOW-L} &\textbf{GLOW-G} &\textbf{GLOW-N} &\textbf{GLOW-GN} \\
\hline
& Arxiv2023        & 1  & 62 & 41& 31       & 4& 59     & 64     & 66     & \textbf{81}      &    \\
& ogbn-arxiv       & 1  & 70 & 59& 12       & 14&      63     & 71     & 60      & \textbf{77} \\
& ogbn-product     & 1  & 29 & 45& 34       & 5& 41     & 44     & 51     & \textbf{57}      &    \\ \cline{2-11}
& drug-superclass  & 1  & 88 & 84& 29       & 3& 50     & 77     & 85     & \textbf{92}      &    \\
& drug-kingdom     & 1  & 98 & 96& 88       & 94& 91     & \textbf{98}     & 97     & 96      &    \\
& protein-SPECIES  & 1  & 75 & 89& 8        & 16& 19     & 71     & 91     & \textbf{92}      &    \\
& protein-FAMILY   & 1  & 67 & 70& 9        & 13& 35     & 39     & 69     & \textbf{71}      &    \\
& film-country     & 1  & 55 & 55& 40       & 4& 40     & \textbf{60}     & 57     & 55      &    \\
& film-subject     & 1  & 77 & 81& 15       & 7& 50     & 30     & \textbf{82}     & 80      &    \\
& film-language    & 1  & \textbf{87} & 83& 70       & 13& 73     & 80     & 85     & 86      &    \\
& person-nationality        & 1  & 86 & 89& 27       & 38& 45     & 89     & \textbf{91}     & 90      &    \\
& parson-graduateOfOrg      & 1  & 51 & 53& 3        & 25& 27     & 36     & 54     & \textbf{55}      &    \\
& person-occupation& 1  & 48 & 46& 7        & 12& 15     & 38     & 46     & \textbf{53}      &    \\
& person-spokenLang& 1  & 86 & 83& 70       & 61& 83     & \textbf{89}     & 83     & 85      &    \\
& person-givenAward& 1  & 93 & 91& 7        & 50& 75     & 93     & 93     & \textbf{93}      &    \\
& CWork-PublishedLang       & 1  & 75 & 80& 50       & 60& 55     & 55     & 85     & \textbf{95}      &    \\
& CWork-country    & 1  & 72 & 51& 64       & 43& 62     & 75     & 48     & \textbf{79}      &    \\
& person-title     & 1  & 79 & 72& 5        & 41& 63     & 47     & 77     & \textbf{83}      &    \\
& person-Company   & 1  & 97 & 95& 3        & 93& 30     & 21     & 96     & \textbf{97}      &    \\
& person-InvestementRegion  & 1  & 98 & 94& 7        & 46& 95     & 87     & 96     & \textbf{100}     &    \\
& person-InvestementCountry & 1  & \textbf{98} & 95& 38       & 14& 94     & 55     & 97     & 97      &    \\ \cline{2-11}
& drug-class       & 2  & 46 & 68& 11       & 15& 33     & 45     & 70     & \textbf{73}      &    \\
& protein-keyword  & 2  & 35 & 42& 5        & 7& 33     & 53     & 44     & \textbf{57}      &    \\
& film-genre       & 2  & 54 & 51& 50       & 25& 60     & \textbf{62}     & 51     & 52      &    \\
& film-producer    & 2  & \textbf{35} & 21& 7        & 7& 14     & 24     & 21     & 28      &    \\
& CWork-ProductionCompany   & 2  & 20 & 22& 10       & 29& 21     & \textbf{31}     & 25     & 28      &    \\
& CWork-Genere     & 2  & 20 & 22& 16       & 7& 22     & \textbf{33}     & 19     & 27      &    \\
\multirow{-28}{*}{\textbf{\makecell{Qwen3:8b}}}& CWork-publisher  & 2  & 31 & 28& 33       & 17& 58     & \textbf{62}     & 27     & 32      &    \\ \hline
\end{tabular}
\end{table}

\definecolor{rowgray}{gray}{0.95}
\begin{table}[!htp]\centering
\rowcolors{2}{rowgray}{white}
  \caption{The detailed accuracy (\%) of {\sysName} compared to baseline models on existing question answering datasets from \cite{Ask-GNN} and our developed benchmark {\Taxonomy}. All results below use GPT-4o-Mini~\cite{gpt40mini} as the underlying LLM. Best results are marked in \textbf{bold}. }
\scriptsize
\begin{tabular}{lcc|ccc|ccccc}\toprule
\label{tab:exp_res_gpt4omini}
LLM-Model &Dataset &RH &AskGNN &GraphSAINT &LLM-Only &\textbf{GLOW-L} &\textbf{GLOW-G} &\textbf{GLOW-N} &\textbf{GLOW-GN} \\
\hline
&Arxiv2023 &1 &N/A &41 &35 &58 &59 &48 &\textbf{62} \\
&ogbn-arxiv &1&N/A&41&35&58&59&48&\textbf{67} \\
&ogbn-product &1 &N/A &45 &N/A &N/A &N/A &N/A &N/A \\
\cline{2-10}
&drug-superclass &1 &N/A &84 &8 &47 &57 &73 &\textbf{87} \\
&drug-kingdom &1 &N/A &96 &31 &\textbf{100} &\textbf{100} &98 &99 \\
&protein-SPECIES &1 &N/A &89 &6 &10 &77 &92 &\textbf{93} \\
&protein-FAMILY &1 &N/A &70 &11 &53 &43 &69 &\textbf{72} \\
&film-country &1 &N/A &55 &47 &47 &\textbf{71} &57 &58 \\
&film-subject &1 &N/A &\textbf{81} &15 &73 &38 &76 &75 \\
&film-language &1 &N/A &83 &80 &80 &80 &86 &\textbf{88} \\
&person-nationality &1 &N/A &89 &43 &56 &89 &\textbf{92} &87 \\
&parson-graduateOfOrg &1 &N/A &53 &19 &22 &33 &55 &\textbf{57} \\
&person-occupation &1 &N/A &46 &7 &23 &30 &46 &\textbf{52} \\
&person-spokenLang &1 &N/A &83 &70 &81 &\textbf{91} &83 &89 \\
&person-givenAward &1 &N/A &91 &6 &68 &\textbf{96} &93 &91 \\
&CWork-PublishedLanguage &1 &N/A &80 &65 &60 &65 &83 &\textbf{87} \\
&CWork-country &1 &N/A &51 &62 &64 &77 &48 &\textbf{83} \\
&person-title &1 &N/A &72 &5 &38 &55 &74 &\textbf{77} \\
&person-company &1 &N/A &95 &3 &6 &30 &96 &\textbf{98} \\
&person-InvestementRegion &1 &N/A &94 &9 &7 &96 &\textbf{100} &96 \\
&person-InvestementCountry &1 &N/A &95 &41 &44 &\textbf{100} &97 &97 \\
\cline{2-10}
&drug-class &2 &N/A &68 &8 &27 &45 &\textbf{69} &67 \\
&protein-keyword &2 &N/A &42 &7 &28 &44 &40 &\textbf{45} \\
&film-genre &2 &N/A &51 &68 &65 &\textbf{70} &50 &53 \\
&film-producer &2 &N/A &21 &20 &17 &20 &21 &\textbf{31} \\
&CWork-ProductionCompany &2 &N/A &22 &10 &28 &\textbf{42} &25 &32 \\
&CWork-Genere &2 &N/A &22 &27 &27 &\textbf{72} &19 &44 \\
\multirow{-28}{*}{\textbf{GPT-4o-Mini}} &CWork-publisher &2 &N/A &28 &62 &70 &\textbf{79} &25 &75 \\
\bottomrule
\end{tabular}
\end{table}
\definecolor{rowgray}{gray}{0.95}
\begin{table}[!htp]\centering
\rowcolors{2}{rowgray}{white}
  \caption{The detailed accuracy (\%) of {\sysName} compared to baseline models on existing question answering datasets from \cite{Ask-GNN} and our developed benchmark {\Taxonomy}. All results below use DeepSeek-V3~\cite{deepseek-llm} as the underlying LLM. Best results are marked in \textbf{bold}. }
\scriptsize
\begin{tabular}{lcc|ccc|ccccc}\toprule
\label{tab:exp_res_DeepSeek-V3}
LLM-Model &Dataset &RH &AskGNN &GraphSAINT &LLM-Only &\textbf{GLOW-L} &\textbf{GLOW-G} &\textbf{GLOW-N} &\textbf{GLOW-GN} \\
\hline
 &Arxiv2023 &1 &N/A &41 &17 &50 &57 &61 &\textbf{73} \\
    &ogbn-arxiv &1 &N/A &59 &58 &60 &62 &63 &\textbf{65} \\
    &ogbn-product &1 &N/A &45 &N/A &N/A &N/A &N/A &N/A \\
    \cline{2-10}
    &drug-superclass &1 &N/A &84 &3 &32 &48 &73 &\textbf{86} \\
    &drug-kingdom &1 &N/A &96 &9 &88 &\textbf{100} &98 &96 \\
    &protein-SPECIES &1 &N/A &89 &2 &8 &79 &92 &\textbf{93} \\
    &protein-FAMILY &1 &N/A &70 &21 &76 &\textbf{87} &69 &81 \\
    &film-country &1 &N/A &55 &57 &65 &70 &57 &\textbf{71} \\
    &film-subject &1 &N/A &81 &23 &80 &38 &80 &\textbf{83} \\
    &film-language &1 &N/A &83 &81 &86 &86 &86 &\textbf{93} \\
    &person-nationality &1 &N/A &89 &72 &81 &87 &91 &\textbf{93} \\
    &parson-graduateOfOrg &1 &N/A &53 &32 &48 &43 &52 &\textbf{54} \\
    &person-occupation &1 &N/A &46 &19 &34 &34 &46 &\textbf{53} \\
    &person-spokenLang &1 &N/A &83 &79 &79 &82 &83 &\textbf{85} \\
    &person-givenAward &1 &N/A &91 &3 &56 &58 &93 &\textbf{94} \\
    &CWork-PublishedLanguage &1 &N/A &80 &47 &50 &50 &84 &\textbf{90} \\
    &CWork-country &1 &N/A &51 &68 &75 &\textbf{87} &48 &79 \\
    &person-title &1 &N/A &72 &41 &41 &63 &74 &\textbf{77} \\
    &person-company &1 &N/A &95 &3 &24 &33 &96 &\textbf{97} \\
    &person-InvestementRegion &1 &N/A &94 &7 &21 &100 &89 &\textbf{100} \\
    &person-InvestementCountry &1 &N/A &95 &44 &58 &100 &97 &\textbf{100} \\
\cline{2-10}
    &drug-class &2 &N/A &68 &7 &21 &39 &69 &\textbf{70} \\
    &protein-keyword &2 &N/A &42 &3 &40 &\textbf{59} &49 &52 \\
    &film-genre &2 &N/A &51 &53 &60 &58 &50 &\textbf{62} \\
    &film-producer &2 &N/A &21 &10 &30 &\textbf{50} &14 &34 \\
    &CWork-ProductionCompany &2 &N/A &22 &28 &53 &\textbf{57} &25 &53 \\
    &CWork-Genere &2 &N/A &22 &18 &22 &27 &16 &\textbf{28} \\
    \multirow{-28}{*}{\textbf{DeepSeek-V3}}&CWork-publisher &2 &N/A &28 &70 &79 &\textbf{87} &16 &70 \\
\bottomrule
\end{tabular}
\end{table}
\definecolor{rowgray}{gray}{0.95}
\begin{table}[!htp]\centering
\rowcolors{2}{rowgray}{white}
  \caption{The detailed accuracy (\%) of {\sysName} compared to baseline models on existing question answering datasets from \cite{Ask-GNN} and our developed benchmark {\Taxonomy}. All results below use DeepSeek-R1-Distill-Qwen:7B~\cite{deepseek-r1-7b} as the underlying LLM. Best results are marked in \textbf{bold}. }
\scriptsize
\begin{tabular}{lcc|cccc|ccccc}\toprule
\label{tab:exp_res_DeepSeek-R1-7B}
LLM-Model &Dataset &RH &AskGNN &GraphSAINT &LLM-Only&GCR &\textbf{GLOW-L} &\textbf{GLOW-G} &\textbf{GLOW-N} &\textbf{GLOW-GN} \\
\hline
    
    & Arxiv2023& 1  & 48& 41    & 24  & 7  & 44  & 55  & 59  & \textbf{67} \\
& ogbn-arxiv    & 1  & 50& 59    & 13  & 15 & 30  & 30  & 61  & \textbf{63} \\
& ogbn-product  & 1  & 19& 45    & 27  & 12 & 36  & 38  & 43  & \textbf{51} \\ \cline{2-11} 
& drug-superclass & 1  & 53& 84    & 20  & 5  & 20  & 21  & \textbf{69}& 68   \\
& drug-kingdom  & 1  & 98& 96    & 60  & 90 & 90  & \textbf{100}    & 98  & \textbf{100}\\
& protein-SPECIES & 1  & 71& 89    & 1   & 12 & 4   & 62  & 88  & \textbf{90} \\
& protein-FAMILY& 1  & 62& 70    & 7   & 23 & 30  & 35  & 68  & \textbf{69} \\
& film-country  & 1  & 56& 55    & 17  & 34 & 44  & 52  & 57  & \textbf{64} \\
& film-subject  & 1  & 78& 81    & 8   & 8  & 11  & 34  & 76  & \textbf{82} \\
& film-language & 1  & 86& 83    & 73  & 32 & 73  & 66  & 86  & \textbf{88} \\
& person-nationality   & 1  & 85& 89    & 22  & 41 & 30  & 79  & \textbf{91}& 89   \\
& parson-graduateOfOrg & 1  & 50& 53    & 5   & 17 & 19  & 36  & 53  & \textbf{54} \\
& person-occupation    & 1  & 47& 46    & 3   & 12 & 15  & 31  & 46  & \textbf{54} \\
& person-spokenLang    & 1  & 82& 83    & 35  & 56 & 41  & 79  & 81  & \textbf{83} \\
& person-givenAward    & 1  & 89& 91    & 8   & 23 & 50  & 70  & 90  & \textbf{92} \\
& CWork-PublishedLanguage   & 1  & 71& 80    & 50  & 44 & 50  & 55  & 77  & \textbf{81} \\
& CWork-country & 1  & 62& 51    & 37  & 42 & 41  & \textbf{70}& 48  & 66   \\
& person-title  & 1  & 73& 72    & 5   & 51 & 44  & 52  & 80  & \textbf{82} \\
& person-company& 1  & 93& 95    & 3   & 90 & 30  & 19  & \textbf{96}& 92   \\
& person-InvestementRegion  & 1  & 92& 94    & 3   & 33 & 64  & 96  & \textbf{100}    & \textbf{100}\\
& person-InvestementCountry & 1  & 93& 95    & 11  & 56 & 72  & 94  & 97  & \textbf{97} \\ \cline{2-11} 
& drug-class    & 2  & 41& 68    & 4   & 13 & 15  & 24  & 69  & \textbf{71} \\
& protein-keyword & 2  & 31& 42    & 3   & 8  & 18  & 22  & 32  & \textbf{38} \\
& film-genre    & 2  & 53& 51    & 35  & 18 & 50  & 56  & 50  & \textbf{57} \\
& film-producer & 2  & 30& 21    & 7   & 9  & 7   & 7   & 14  & \textbf{24} \\
& CWork-ProductionCompany   & 2  & 18& 22    & 10  & 23 & 17  & 21  & 25  & \textbf{32} \\
& CWork-Genere  & 2  & 17& 22    & 11  & 17 & 11  & 16  & 16  & \textbf{27} \\
\multirow{-28}{*}{\textbf{\makecell{DeepSeek-R1-\\Distill-Qwen:7B}}}& CWork-publisher & 2  & 25& 28    & 12  & 23 & 20  & \textbf{40}& 16  & 26   \\
\bottomrule
\end{tabular}
\end{table}
\definecolor{rowgray}{gray}{0.95}
\begin{table}[!htp]\centering
\rowcolors{2}{rowgray}{white}
\caption{The detailed accuracy (\%) of {\sysName} compared to baseline models on existing question answering datasets from \cite{Ask-GNN} and our developed benchmark {\Taxonomy}. All results below use IBM Granite-3.3-8B-instruct~\cite{Granite-8b} as the underlying LLM. Best results are marked in \textbf{bold}. }
\scriptsize
\begin{tabular}{lcc|cccc|cccc}\toprule
\label{tab:exp_res_Granite3.3}
LLM-Model &Dataset &RH &AskGNN &GraphSAINT &LLM-Only &GCR &\textbf{GLOW-G} &\textbf{GLOW-N} &\textbf{GLOW-GN} \\
\hline
& Arxiv2& 1& 63& 41 & 14& 14& 32& 63& \textbf{77}\\
& ogbn-arxiv& 1& 66& 59 & 5& 21& \textbf{31} & 55& \textbf{73}\\
& ogbn-product& 1& 30& 45 & 22& 13& 45& 52& \textbf{54}\\ \cline{2-11} 
& drug-superclass & 1& 86& 84 & 0& 9 & 22 & 73& \textbf{71} \\
& drug-kingdom& 1& 97& 96 & 8 & 98& 95& 98 & \textbf{99} \\
& protein-SPECIES & 1& 75& 89 & 2 & 18& 45 & 93 & \textbf{90} \\
& protein-FAMILY& 1& 67& 70 & 0& 33& 36& 69& \textbf{72} \\
& film-country& 1& 56 & 56& 48 & 60& \textbf{66}& 58 & 55\\
& film-subject& 1& 76& 81 & 12& 11& 27 & 81 & \textbf{85} \\
& film-language & 1& 84& 83 & 87& 46& 80& 87& \textbf{88} \\
& person-nationality& 1& 86& 89& 52& 88& 79& 91 & \textbf{92} \\
& parson-graduateOfOrg& 1& 52& 53& 6 & 14& 31 & 56& \textbf{57} \\
& person-occupation & 1& 47& 46& 8 & 23& 23& 46& \textbf{50}\\
& person-spokenLang & 1& 85& 83& 48& 75& 52& 83& \textbf{88} \\
& person-givenAward & 1& 93& 92& 3& 25& 63& 91 & \textbf{94} \\
& CWork-PublishedLanguage & 1& 81& 80 & 45& 35& 80& 81& \textbf{94}\\
& CWork-country & 1& 72& 51 & 58& 69& \textbf{75} & 59 & \textbf{69} \\
& person-title& 1& 78& 72& 14& 55& 64 & 80& \textbf{83} \\
& person-company& 1& 97& 96& 3& 96& 42 & 96 & \textbf{97} \\
& person-InvestementRegion& 1& 99& 95& 4 & 29& 82& 98& \textbf{100} \\ 
\cline{2-11} 
& person-InvestementCountry & 1& \textbf{99} & 96& 39& 69& 78 & 96& 97\\
& drug-class& 2& 47& 68 & 0& 18& 27& 70& \textbf{70}\\
& protein-keyword & 2& 35& 42 & 2 & 10& 36 & 9& \textbf{38} \\
& film-genre& 2& 54& 51 & 35& 15& 52 & 54& \textbf{57} \\
& film-producer & 2& \textbf{35} &21& 5& 5 & 30& 23& 29\\
& CWork-ProductionCompany & 2& 22&22& 7& 18& 21 & 25& \textbf{32} \\
& CWork-Genere& 2& 21& 22& 16& 22& \textbf{50} & 17 & 33\\
\multirow{-28}{*}{\textbf{\makecell{Granite-3.3:8B-\\Instruct}}}& CWork-publisher & 2& 32.7& 28& 33& 46& \textbf{58} & 27 & 36 \\
\bottomrule
\end{tabular}
\end{table}
\end{document}